\RequirePackage{snapshot}
\documentclass{article}

\usepackage{mycustom}

\title{Variational Predictive Information Bottleneck}

\author{%
  Alexander A.~Alemi \\
  Google Research\\
  \texttt{alemi@google.com}\\
}

\date{\today}

\begin{document}

\maketitle

\begin{abstract}
    In classic papers,~\citet{zellner, zellner2} 
    demonstrated that Bayesian inference could be derived
    as the solution to an information theoretic functional.
    Below we \emph{derive} a generalized form of this functional
    as a variational lower bound of a predictive
    information bottleneck objective.
    This generalized functional encompasses most
    modern inference procedures and suggests novel ones.
\end{abstract}

\section{Introduction}

Consider a data generating process $\phi \sim p(\phi)$ from which we have
some $N$ draws that constitute our training set, $\xp = \{ x_1, x_2, \dots, x_N \} \sim p(x|\phi)$. 
We can also imagine (potentially infinitely many) 
future draws from this same process $\xf = \{ x_{N+1}, \dots \} \sim p(x|\phi)$.  
The \emph{predictive information} $I(\xp; \xf)$\footnote{
We use $I(x; y)$ for the \emph{mutual information} between
two random variables: $I(x; y) \equiv \mathbb{E}_{p(x,y)}\left[ \log\frac{p(x,y)}{p(x)p(y)}\right]$} gives a unique
measure of the complexity of a data generating process~\citep{predictive}.

\begin{figure}[htb]
    \centering
    \begin{tikzpicture}[minimum size=1cm]
\def\r{0.5 cm}
\node [draw, circle] (X) at (0, 0) {$\xp$};
\node [draw, circle] (Theta) at (3*\r, 0) {$\theta$};
\node [draw, circle] (Phi) at (-3*\r, 0) {$\phi$};
\node [draw,circle] (Xf) at (-6*\r,0) {$\xf$};
\draw [->] (X) -- (Theta);
\draw [->] (Phi) -- (X);
\draw [->] (Phi) -- (Xf);
\draw [thick] (-1.5*\r, -1.5*\r) rectangle (1.5*\r, 1.5*\r);
\draw [thick] (-7.5*\r, -1.5*\r) rectangle (-4.5*\r, 1.5*\r);
\end{tikzpicture}
    \caption{
        \label{fig:worldp}
        Graphical model under consideration.
    }
\end{figure}
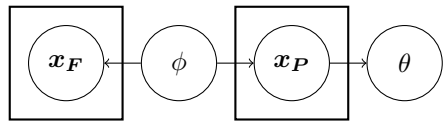

The goal of \emph{learning} is to capture this complexity.
To perform learning, we form a 
global representation of the 
dataset $p(\theta | \xp)$.
This can be thought of as a 
learning algorithm, that,
given a set of observations,
produces a summary statistic of the dataset
that we hope is useful for predicting future draws
from the same process.
This algorithm could be deterministic or more generally, stochastic.  

For example, imagine training a neural network on some data with 
stochastic gradient descent.
Here the training data would be $\xp$, the test data $\xf$ and the 
neural network parameters would be $\theta$. Our training procedure
implicitly samples from the distribution $p(\theta|\xp)$. 

How do we judge the utility
of this learned global representation?
The mutual information $I(\theta; \xf)$ quantifies
the amount of information our representation captures
about future draws.\footnote{
It is interesting to note that in the limit of an infinite number of future draws, $I(\theta; \xf)$ approaches
$I(\theta; \phi)$.
Therefore, the amount of information we have about
an infinite number of future draws from the process is the same as the amount of information
we have about the nature and identity of the data generating process itself.}
To maximize learning we therefore aim to 
maximize this quantity.

This is, of course, only interesting if we
constrain how expressive our global representation is,
for otherwise we could simply retain the full dataset.
The amount of information retained about the observed data: $I(\theta;\xp)$ is a direct measure of our representation's
complexity. The bits a learner extracts from data provides upper bounds on  generalization~\citep{learners}.

\section{Predictive Information Bottleneck}
Combined, these motivate
the \emph{predictive information bottleneck objective}, 
a generalized \emph{information bottleneck}~\citep{predictive,ib}:
\begin{equation}
    \max_{p(\theta|\xp)} I(\theta; \xf) \quad \textrm{ s.t. } \quad I(\theta; \xp) = I_0.
    \label{eqn:pi}
\end{equation}
We can turn this into 
an unconstrained optimization problem with the use of 
a Lagrange multiplier $\beta$:
\begin{equation}
    \max_{p(\theta|\xp)} I(\theta; \xf) - (1-\beta) I(\theta; \xp).
\end{equation}

While this objective seems wholly out of reach, we can
make progress by noting that our random variables satisfy
the Markov chain: $\xf \leftarrow \phi \rightarrow \xp \rightarrow \theta$, in which $\theta$ and $\xf$ are 
conditionally independent given $\xp$:
\begin{equation}
    I(\theta; \xf, \xp) = I(\theta; \xf ) + I(\theta; \xp | \xf) = I(\theta; \xp) + \cancel{I(\theta; \xf|\xp)}.
\end{equation}
This implies:
\begin{equation}
    I(\theta; \xf) = I(\theta; \xp) - I(\theta; \xp | \xf).
\end{equation}
and the equivalent unconstrained optimization
problem:\footnote{
A similar transformation for the (local) variational information bottleneck appeared in \citet{ceb}.}
\begin{equation}
    \min_{p(\theta|\xp)} I(\theta; \xp|\xf) - \beta I(\theta; \xp) .
    \label{eqn:pib}
\end{equation}
The first term here: $I(\theta; \xp | \xf)$ is the residual information between our global 
representation and the dataset after we condition on full knowledge of the data generating procedure.
This is a direct measure of the inefficiency of our proposed representation.

\section{Variational Predictive Information Bottleneck}

Simple variational bounds~\citep{vmibounds} can be derived for this
objective, just as was done for the 
(local) information bottleneck objective in \citet{vib}.
First, we demonstrate a variational upper bound on $I(\theta; \xp | \xf)$:~\footnote{$\langle \cdot \rangle$ is used to denote expectations, and unless denoted otherwise with respect to the full joint density $p(\theta|\xp)p(\xp|\phi)p(\phi)p(\xf|\phi)$}
\begin{equation}
    I(\theta; \xp| \xf) = \left\langle \log \frac{p(\theta|\xp)}{p(\theta|\xf)} \right\rangle
    \leq \left\langle \log \frac{p(\theta | \xp)}{q(\theta)} \right\rangle  .
\end{equation}
Here we upper bound the residual information by using a variational approximation to
$p(\theta | \xf)$, the marginal of our global representation over all datasets drawn from the 
same data generating procedure.  Any distribution $q(\theta)$ independent of $\xf$ suffices.

Next we variationally lower bound $I(\theta; \xp)$ with:
\begin{equation}
   I(\theta; \xp) = \left\langle \log \frac{p(\xp|\theta)}{p(\xp)} \right\rangle 
   \geq H(\xp) + \sum_i \left\langle \log q(x_i|\theta)\right\rangle.
\end{equation}
The entropy of the training data $H(\xp)$ is a constant 
outside of our control that can be ignored. 
Here we variationally approximate the ``posterior''
of our global representation with a factorized 
``likelihood'':
$\prod_i q(x_i|\theta) = q(\xp|\theta) \sim p(\xp|\theta)$.
Notice that while $p(\xp|\theta)$
will not factorize in general, 
we can certainly consider a family of 
variational approximations
that do.

Combining these variational bounds, we generate the objective:
\begin{equation}
    \min_{p(\theta|\xp)} \left\langle \log \frac{p(\theta|\xp)}{q(\theta)} 
    - \beta \sum_i \log q(x_i|\theta) \right\rangle .
    \label{eqn:obj}
\end{equation}
We have thus \emph{derived}, as a variational lower bound on
the predictive information bottleneck, the objective
\citet{zellner} postulates (with $\beta=1$)
is satisfied for inference procedures
that optimally process information.
As \citet{generalizedvi} demonstrates, this
encompasses a wide array of modern inference procedures,
including Generalized Bayesian Inference~\citep{generalbayes}
and a generalized Variational Inference, 
dubbed Gibbs VI~\citep{gibbsvi,gibbsvi2}.\footnote{
To incorporate the Generalized VI~\citep{generalizedvi} with
divergence measures other than KL, we need only replace
our mutual informations (which are KL based) with
their corresponding generalizations.} Below we highlight
some of these and other connections.

\section{Connections}

If, in \Cref{eqn:obj},
we identity $q(\theta)$ with a 
\emph{fixed prior} and $q(x|\theta)$ with a
\emph{fixed likelihood} of a generative model, optimizing
this objective for $p(\theta | \xp)$ in the space
of all probability densities gives the
generalized Boltzmann distribution~\citep{jaynes}:
\begin{equation}
    p(\theta|\xp) = q(\theta) \left[ \prod_i q(x_i|\theta) \right]^{\beta} / \mathcal{Z},
    \label{eqn:sol}
\end{equation}
where $\mathcal{Z}$ is the \emph{partition function}.~\footnote{
$\mathcal{Z} \equiv \int d\theta\, q(\theta) \left[ \prod_i q(x_i|\theta) \right]^\beta $} 
This is a generalized form of Bayesian Inference called
the
\emph{power likelihood}~\citep{powerlikelihood2,powerlikelihood}.
Here the inverse temperature $\beta$ acts as a Lagrange multiplier
controlling the trade-off between the amount of information we retain about our observed data ($I(\theta; \xp)$) and how much predictive information we capture ($I(\theta; \xf)$).  
As $\beta \to \infty$ (temperature goes to zero), we recover
the maximum likelihood solution. 
At $\beta = 1$ (temperature = 1) we recover ordinary Bayesian inference. 
As $\beta \to 0$ (temperature goes to infinity), we recover just prior
predictive inference that ignores the data entirely. 
These limits are summarized in \Cref{tab:results}.

\begin{table}[htbp]
\begin{center}
\begin{adjustbox}{width={\textwidth},totalheight={\textheight},keepaspectratio}%
\begin{tabular}{ c | c | c | c }
  Limit & Inference & Equivalent Objective & $p(\theta|\xp)$ \\ 
  \hline
  $\beta$ & Generalized Bayes & $\max I(\theta; \xf) - (1-\beta) I(\theta; \xp)$ & $\propto q(\theta) q(\xp|\theta)^\beta$ \\
  $\beta \to 0$ & Prior Predictive & $\min I(\theta; \xp | \xf)$ & $q(\theta)$ \\
  $\beta \to 1$ & Bayesian & $\max I(\theta; \xf)$ & $\propto q(\theta) q(\xp|\theta)$ \\
  $\beta \to \infty$ & Maximum Likelihood & $\max I(\theta; \xp)$ & $\arg\min_\theta q(\xp|\theta)$ \\
\end{tabular}
\end{adjustbox}
\caption{
    Power Bayes can be recovered as a variational lower bound on the predictive information bottleneck
    objective~(\Cref{eqn:pib}).
    \label{tab:results}
}
\end{center}
\end{table}

More generally, notice that in~\Cref{eqn:obj} the densities
$q(x|\theta)$ and $q(\theta)$ 
are not literally the likelihood and prior of a generative model,
they are variational approximations that we
have complete freedom to specify.
This allows us to describe other
more generalized forms of Bayesian inference
such as Divergence Bayes or the full Generalized Bayes~\citep{generalizedvi,generalbayes}
provided we can interpret the chosen loss function as 
a conditional distribution.

If we limit the domain of $p(\theta | \xp)$ to a restricted
family of parametric distributions,
we immediately recover not only
standard variational inference, but a broad generalization
known as Gibbs Variational Inference~\citep{generalizedvi,gibbsvi,gibbsvi2}.


Furthermore, nothing prevents us from making $q(x|\theta)$ or $q(\theta)$ themselves parametric
and simultaneously optimizing those.
Optimizing the prior 
with a fixed likelihood,
unconstrained $p(\theta|\xp)$,
and $\beta=1$ the
objective mirrors
Empirical Bayesian~\citep{empiricalbayes}
approaches, including  the notion of 
\emph{reference priors}~\citep{rationalignorance,reference}.
Alternatively,
optimizing a parametric likelihood
with 
a parametric representation $p(\theta | \xp)$,
fixed prior,
and $\beta=1$
equates to a Neural Process~\citep{neuralprocesses}.

%

Consider next \emph{data augmentation}, where we have some stochastic process that modifies
our data with implicit conditional density $t(x'|x)$.  If the augmentation procedure is 
centered about zero so that $\langle x' \rangle_{t(x'|x)} = x$ and our chosen likelihood function is
concave, then we have:
\begin{equation}
    \log q(x|\theta) = \log q(\langle x' \rangle_{x'\sim t(x'|x)} | \theta) \geq \left\langle \log q(x'|\theta) \right\rangle_{x' \sim t(x'|x)},
\end{equation}
which maintains our bound.  For example, for an exponential family likelihood 
and any centered augmentation procedure (like additive mean zero noise), 
doing generalized Bayesian inference on an augmented dataset is also a lower
bound on the predictive information bottleneck objective.

\section{Conclusion and Future Work}

We have shown that a wide range of
existing inference techniques are 
variational lower bounds on
a single predictive information bottleneck objective.

This connection highlights the 
drawbacks of these traditional forms of inference.
In all cases considered in the previous section,
we made two choices
that loosened our variational bounds. First,
we approximated $p(\xp|\theta)$, with a factorized
approximation $q(\xp|\theta) = \prod_i q(x_i|\theta)$.
Second, we approximated the future conditional 
\emph{marginal} $p(\theta|\xf) = \int d\xp\, p(\theta|\xp)p(\xp|\xf)$ as an unconditional ``prior''.
Neither of these approximations is necessary.

For example, consider the following tighter ``prior'':
\begin{equation}
    q(\theta | \xf) \sim \int d\xp' \, p(\theta | \xp') q(\xp' | \xf).
\end{equation}
Here we reuse a tractable global
representation $p(\theta|\xp)$ and instead 
create a variational approximation to the density of alternative
datasets drawn from the same process: $q(\xp'|\xf)$.

We believe this information-theoretic, 
representation-first perspective on learning
has the potential to motivate new 
and better forms of inference.~\footnote{Extensions of this representation-first perspective
beyond a single global representation
towards supervised or unsupervised learning
can be found in~\citet{therml}.}

\bibliographystyle{plainnat}
\bibliography{bib}

\begin{thebibliography}{20}
\providecommand{\natexlab}[1]{#1}
\providecommand{\url}[1]{\texttt{#1}}
\expandafter\ifx\csname urlstyle\endcsname\relax
  \providecommand{\doi}[1]{doi: #1}\else
  \providecommand{\doi}{doi: \begingroup \urlstyle{rm}\Url}\fi

\bibitem[Alemi and Fischer(2018)]{therml}
Alexander~A. Alemi and Ian Fischer.
\newblock {TherML}: Thermodynamics of machine learning, 2018.

\bibitem[Alemi et~al.(2016)Alemi, Fischer, Dillon, and Murphy]{vib}
Alexander~A Alemi, Ian Fischer, Joshua~V Dillon, and Kevin Murphy.
\newblock Deep variational information bottleneck.
\newblock \emph{arXiv preprint arXiv:1612.00410}, 2016.

\bibitem[Alquier et~al.(2016)Alquier, Ridgway, and Chopin]{gibbsvi}
Pierre Alquier, James Ridgway, and Nicolas Chopin.
\newblock On the properties of variational approximations of gibbs posteriors.
\newblock \emph{Journal of Machine Learning Research}, 17\penalty0
  (236):\penalty0 1--41, 2016.
\newblock URL \url{http://jmlr.org/papers/v17/15-290.html}.

\bibitem[Bassily et~al.(2017)Bassily, Moran, Nachum, Shafer, and
  Yehudayoff]{learners}
Raef Bassily, Shay Moran, Ido Nachum, Jonathan Shafer, and Amir Yehudayoff.
\newblock Learners that use little information.
\newblock \emph{arXiv preprint arXiv:1710.05233}, 2017.

\bibitem[Berger et~al.(2009)Berger, Bernardo, Sun, et~al.]{reference}
James~O Berger, Jos{\'e}~M Bernardo, Dongchu Sun, et~al.
\newblock The formal definition of reference priors.
\newblock \emph{The Annals of Statistics}, 37\penalty0 (2):\penalty0 905--938,
  2009.

\bibitem[Bialek et~al.(2001)Bialek, Nemenman, and Tishby]{predictive}
William Bialek, Ilya Nemenman, and Naftali Tishby.
\newblock Predictability, complexity, and learning.
\newblock \emph{Neural computation}, 13\penalty0 (11):\penalty0 2409--2463,
  2001.

\bibitem[Bissiri et~al.(2016)Bissiri, Holmes, and Walker]{generalbayes}
P.~G. Bissiri, C.~C. Holmes, and S.~G. Walker.
\newblock A general framework for updating belief distributions.
\newblock \emph{Journal of the Royal Statistical Society: Series B (Statistical
  Methodology)}, 78\penalty0 (5):\penalty0 1103--1130, 2016.
\newblock \doi{10.1111/rssb.12158}.
\newblock URL
  \url{https://rss.onlinelibrary.wiley.com/doi/abs/10.1111/rssb.12158}.

\bibitem[Fischer(2019)]{ceb}
Ian Fischer.
\newblock The conditional entropy bottleneck, 2019.
\newblock URL \url{https://openreview.net/forum?id=rkVOXhAqY7}.

\bibitem[Futami et~al.(2017)Futami, Sato, and Sugiyama]{gibbsvi2}
Futoshi Futami, Issei Sato, and Masashi Sugiyama.
\newblock Variational inference based on robust divergences.
\newblock \emph{arXiv preprint arXiv:1710.06595}, 2017.

\bibitem[Garnelo et~al.(2018)Garnelo, Schwarz, Rosenbaum, Viola, Rezende,
  Eslami, and Teh]{neuralprocesses}
Marta Garnelo, Jonathan Schwarz, Dan Rosenbaum, Fabio Viola, Danilo~J Rezende,
  SM~Eslami, and Yee~Whye Teh.
\newblock Neural processes.
\newblock \emph{arXiv preprint arXiv:1807.01622}, 2018.

\bibitem[Holmes and Walker(2017)]{powerlikelihood2}
CC~Holmes and SG~Walker.
\newblock Assigning a value to a power likelihood in a general bayesian model.
\newblock \emph{Biometrika}, 104\penalty0 (2):\penalty0 497--503, 2017.

\bibitem[Jaynes(1957)]{jaynes}
Edwin~T Jaynes.
\newblock Information theory and statistical mechanics.
\newblock \emph{Physical review}, 106\penalty0 (4):\penalty0 620, 1957.

\bibitem[Knoblauch et~al.(2019)Knoblauch, Jewson, and Damoulas]{generalizedvi}
Jeremias Knoblauch, Jack Jewson, and Theodoros Damoulas.
\newblock Generalized variational inference.
\newblock \emph{arXiv preprint arXiv:1904.02063}, 2019.

\bibitem[Maritz and Lwin(2018)]{empiricalbayes}
Johannes~S Maritz and T~Lwin.
\newblock \emph{Empirical bayes methods}.
\newblock Routledge, 2018.

\bibitem[Mattingly et~al.(2018)Mattingly, Transtrum, Abbott, and
  Machta]{rationalignorance}
Henry~H Mattingly, Mark~K Transtrum, Michael~C Abbott, and Benjamin~B Machta.
\newblock Maximizing the information learned from finite data selects a simple
  model.
\newblock \emph{Proceedings of the National Academy of Sciences}, 115\penalty0
  (8):\penalty0 1760--1765, 2018.

\bibitem[Poole et~al.(2019)Poole, Ozair, Oord, Alemi, and Tucker]{vmibounds}
Ben Poole, Sherjil Ozair, Aaron van~den Oord, Alexander~A Alemi, and George
  Tucker.
\newblock On variational bounds of mutual information.
\newblock \emph{arXiv preprint arXiv:1905.06922}, 2019.

\bibitem[Royall and Tsou(2003)]{powerlikelihood}
Richard Royall and Tsung-Shan Tsou.
\newblock Interpreting statistical evidence by using imperfect models: robust
  adjusted likelihood functions.
\newblock \emph{Journal of the Royal Statistical Society: Series B (Statistical
  Methodology)}, 65\penalty0 (2):\penalty0 391--404, 2003.

\bibitem[Tishby et~al.(2000)Tishby, Pereira, and Bialek]{ib}
Naftali Tishby, Fernando~C Pereira, and William Bialek.
\newblock The information bottleneck method.
\newblock \emph{arXiv preprint physics/0004057}, 2000.

\bibitem[Zellner(1988)]{zellner}
Arnold Zellner.
\newblock Optimal information processing and bayes's theorem.
\newblock \emph{The American Statistician}, 42\penalty0 (4):\penalty0 278--280,
  1988.

\bibitem[Zellner(2002)]{zellner2}
Arnold Zellner.
\newblock Information processing and bayesian analysis.
\newblock \emph{Journal of Econometrics}, 107\penalty0 (1-2):\penalty0 41--50,
  2002.

\end{thebibliography}

\end{document}